\documentclass[twocolumn]{article}

\usepackage{amsthm,amsmath}
\usepackage{url}
\usepackage{subfig}
\usepackage{multirow}
\usepackage{here}

\usepackage{graphicx}

\title{Leaf segmentation through the classification of edges}

\author{Jonathan Bell \and Hannah M. Dee \thanks{Department of Computer Science, Aberystwyth University, Aberystwtyh SY23 3DB UK}}

\begin{document}

\maketitle

\begin{abstract} % abstract
We present an approach to leaf level segmentation of images of
Arabidopsis thaliana plants based upon detected edges. We introduce 
a novel approach to edge classification, which forms an important part of a method 
to both count the leaves and 
establish the leaf area of a growing plant from images obtained in a 
high-throughput phenotyping system. Our technique
uses a relatively shallow convolutional neural network to classify image edges
as background, plant edge, leaf-on-leaf edge or internal leaf noise. The edges themselves were
found using the Canny edge detector and the classified edges can be used with 
simple image processing techniques to generate a region-based segmentation in which
the leaves are distinct. This approach is strong at distinguishing
occluding pairs of leaves where one leaf is largely hidden, a situation which 
has proved troublesome for plant image analysis systems in the past. 
In addition, we introduce the publicly available plant image dataset that was used for this work.
\end{abstract}

\section{Introduction} \label{sec:intro}
One current growth area in computer vision and in plant science 
is the use of high volume image
capture to obtain measurements of plant growth. Phenotyping installations
such as the National Plant Phenomics Centre in Aberystwyth are capable of
automatically recording images of hundreds of plants, at a temporal frequency which
has previously been difficult. This leads to a new set of problems and 
challenges associated with the extraction of meaningful biological measurements
from image data. 
Computer vision 
techniques show one way of easing the phenotyping bottleneck
%\emph{Arabidopsis thaliana} (Arabidopsis) is an important plant in this area, 
%it was the first plant to have its genome sequenced and is small and fast to grow to maturity.
as they enable measurements to be obtained from an image of a plant automatically. 
Image-based measurements can be gathered without harvesting the plant, 
so a single specimen can provide data throughout its life. 
This not only makes more data available, it also allows acquisition of new types of
data, such as fine-grained variations in growth rate.

We present an image analysis technique that can be used to automate the extraction of
leaf level measurements of plants, to increase the richness of the data 
available.
Image analysis techniques which report projected leaf area are well developed, 
enabling the separation the plant from the background e.g. \cite{DeVylder2011a}. 
These measures allow approximation of plant-level characteristics such as biomass, 
but the relationship breaks down under leaf-on-leaf occlusion seen in older
plants. More subtle measurements, such as number of 
leaves and
the area of each leaf provide more useful information, and it is in this area that the
current paper makes its contributions. We developed our approach using images
of Arabidopsis thaliana (Arabidopsis) captured using a high throughput 
plant phenotyping platform. As Arabidopsis is a rosette plant whose leaves 
overlap as they grow, it is important
to identify regions where this overlap tales place. The dataset captured
to enable this work is also made available and is introduced in this paper 
alongside a description of our image analysis technique.

The background to this work is described in the next section and this is 
followed by a description of the dataset we used for our work in Section~\ref{sec:dataset}. Our methods are  
described in Section~\ref{sec:method}, followed by a description of our results in Section~\ref{sec:results}.
A discussion of these results and the contribution of the current work are presented
in Section~\ref{sec:discussion} and we conclude in Section~\ref{sec:conclusion}.

\section{Background} \label{sec:background}

The use of image analysis techniques in plant phenotyping is of interest both to biologists, who need ways of managing the masses of data generated by high throughput
phenotyping platforms \cite{Furbank2011a}, and computer vision researchers such as
ourselves for whom plant imaging is a challenging new problem domain. Much
plant measurement to date has been destructive (e.g. biomass recorded after harvest) thus imaging opens up the possibility
of providing \emph{more} data as measurements from the same plant can can be taken 
repeatedly over time. 
This means that dynamic differences between plants, such as rates of growth, 
will not be lost as they would be if all measurements are taken at the 
end of an experiment \cite{Dhondt2014a}. In our work we concentrate on 
Arabidopsis. This is
a good plant for image analysis research as it is small, quick to grow and measurements 
can be taken from top down images, given suitable analysis techniques. As the 
first plant to have its genome sequenced it is of major interest to biological
scientists too.
Image analysis techniques have been used in studying Arabidopsis since 1992
\cite{Engelmann1992a} but here we confine our review to approaches to the problem of
leaf level segmentation of
the plant. Readers seeking a broader overview of image analysis and
Arabidopsis will find one in \cite{Bell2016b}.

Overall plant area (projected leaf area) is a good initial predictor of a plant's 
weight \cite{Leister1999a} but as the plant grows and leaves
occlude each other obscuring biomass from view, this relationship changes. 
Leaf level analysis offers the possibility of estimating biomass 
 more accurately, particularly if the technique can deal with heavy occlusion.
Leaf count is also important as the number of leaves is itself an indicator of the plant's maturity 
\cite{Boyes2001a}. There are therefore advantages to approaches that are
capable of doing more than simply separating the plant from the background (so called
``whole-plant'' techniques), that is, whole-plant imaging could well mask differences
between different genotypes. Leaf-level segmentation could, for example, allow 
rates of growth of individual leaves to be measured from time lapse sequences 
of images \cite{Minervini2015a}.  Interest in this area in the computer vision community has been 
stimulated by the Leaf Counting and Leaf Segmentation Challenges
\cite{Minervini2015c, Scharr2014a} and by the release of annotated 
datasets such as the Leaf Segmentation Challenge sets and 
MSU-PID \cite{Cruz2015a}. 

The Leaf Segmentation Challenge 2015 entries are collated in
\cite{Scharr2015a}, providing a useful introduction to state-of-the-art
techniques for leaf-level image analysis.  Pape and Klukas \cite{Pape2014a} used
3D histograms for pixel classification, building on this by
comparing various machine learning approaches \cite{Pape2015a} using 
classifiers provided by Weka. Like us, Pape and Klukas have
concentrated some of their efforts on the robust detection of leaf edges. They do not report an overall best method, and indeed found
that each of the the three sets of 
Leaf Segmentation Challenge data (two of Arabidopsis and one of tobacco) respond best to a different machine learning technique.
Vukadinovic and Polder
\cite{Vukadinovic2015a} use neural network based pixel classification to sort plant from 
background and a watershed approach to find individual leaves. Seeds for 
these watersheds were found using Euclidean distance from the previously 
determined background mask. Yin et al. \cite{Yin2014a} use chamfer matching to match leaves with one of the set of possible templates.
This approach was originally developed for chlorophyll fluorescence
images, and developed in a tracking context to make use of time lapse data
by using the result of the previous image analysis 
in place of image templates for analysing the next image \cite{Yin2014b}.  As described in \cite{Scharr2015a}, a team from Nottingham University have used SLIC superpixels
for both colour-based plant segmentation and watershed based leaf segmentation. All the work described in \cite{Scharr2015a} was tested using
the three sets of data in the Leaf Segmentation Challenge set, using
separate subsets of each set for training and test data, so results quoted
in \cite{Scharr2015a} are for methods tailored (as far as possible) for that data set.

A recent development in image analysis for plant phenotyping is the
exploration of deep learning approaches (see LeCun et al 
\cite{LeCun2015a} for a general introduction to this field).
Such approaches are often used for object classification, and Pound et al. \cite{Pound2016a} have demonstrated promising results for
plant phenotyping work, using Convolutional Neural Networks (CNNs) to
identify the presence of root end and features such as tips and bases
of leaves and ears of cereals. Closer to the present work, both
Romero-Paredes and Torr \cite{Romera-Paredes2016a} and Ren and Zemel
\cite{Ren2017} have separately achieved state of the art results for leaf segmentation and counting using Recurrent Neural Networks, which are capable of keeping track of objects that have previously been identified.
Both used the leaf segmentation challenge dataset, and both use the leaf, not the plant as the object to be counted.

\section {Arabidopsis plant image dataset} \label{sec:dataset}

As part of this work we have generated a dataset of several thousand top
down images of growing Arabidopsis plants. 
The data has been made available to the wider community as the Aberystwyth Leaf Evaluation Dataset (ALED)
\footnote{The dataset is hosted at \url{https://zenodo.org} and it can be found from
\url{https://doi.org/10.5281/zenodo.168158}}.
The images were obtained
using the Photon Systems Instruments (PSI) PlantScreen plant scanner 
\cite{PSIweb} at the National Plant Phenomics Centre.
Alongside the images themselves are some ground truth hand-annotations and
software to evaluate leaf level segmentation against this ground truth.

\subsection{Plant material} \label{sec:dataset:plant}
The plant material imaged in this dataset is Arabidopsis thaliana, 
accession Columbia (Col-0). Plants were grown in 
180ml compost (Levington F2 + 205 grit sand) in PSI 6cm square pots.
The plants were grown with gravimetric watering to 75\% field capacity
under a 10 hour day 20$^\circ$C/15$^\circ$C glasshouse regime on the PSI platform. 
Plants were sown 23 November 2015 and pricked out into weighed
pots at 10 days old. 80 plants were split between four trays.
Two plants were removed from each tray at 30, 37, 43, 49 and 53
days after sowing and final harvest was 56 days after sowing.
Harvested plants were sacrificed to take physical measurements such
as dry weight, wet weight, and leaf count.
The plants were removed in a pattern reducing the density of 
population of each tray to lessen the possibility of 
neighbouring plants occluding each other. However, there are 
instances where this has happened towards the end of the dataset.

\subsection{Plant images} \label{sec:dataset:images}
The plants were photographed from above (``top view'') using the visible 
spectrum approximately every 13 minutes 
during the 12 hour daylight period, starting on 14 December 2015, 21 days after 
sowing.  There are some short gaps in the image sequence 
attributable to machine malfunctions.
The imaging resulted in the acquisition of 1676 images of each tray, each image
having 20, 18, 16, 14, 12 or 10 plants as plants were harvested.
Images were taken using the PSI platform's built in camera,
an IDS uEye 5480SE, resolution 2560*1920 (5Mpx) fitted with a
Tamron 8mm f1.4 lens. This lens exhibits some barrel distortion.
Images were saved as .png files (using lossless compression) with filenames
that incorporate tray number (031, 032, 033 and 034) and date and time of capture. 
Times are slightly different between trays as images were taken sequentially.
Other than png compression, no post-processing was done, which 
means that images have the camera's barrel distortion.
Code to correct this is supplied along with the dataset.

\subsection{Ground truth annotations} \label{sec:dataset:annotations}
Our dataset has accompanying ground truth annotations of the last image of each day 
of one tray (number 31), together with the first image taken 21, 28, 35 and 42 
days after sowing and the first image taken after 2pm 
28, 35, 42 and 49 days after sowing. 
This amounts to 43 images with between 10 and 20 plants in each image 
and a total of 706 annotated plant images from tray 031, as shown in the ``\emph{train}'' columns of Table 
\ref{table:gt}.
The suggestion is that these are used as training data.
We also have ground truth annotations of the first image taken of tray 032 and the 
first image taken after 2pm every third day, starting 22 days after sowing. 
This adds another 210 plant annotations shown in the ``\emph{test}'' columns of Table \ref{table:gt}.
Ground truth was created through hand annotations of individual leaves, using a drawing package. This
annotation was carried out by casually employed postgraduate students.

\begin{table}
\centering
\caption{Images in the suggested training and test data sets along with days after sowing information and number of individual plants}
\label{table:gt}
\begin{tabular}{p{0.9cm} | p{0.95 cm} | p{1cm} | p{1cm} | p{1cm} | p{1cm} }
Plant count & D.A.S. & Images (train) & Plants (train) &Images (test) & Plants (test) \\
\hline
20 & 21 22 25 28 & 12 & 240 & 4 & 80 \\
18 & 31 34 & 9 & 162 & 2 & 36 \\
16 & 37 40 & 8 & 120 & 2 & 32 \\
14 & 43 46 & 7 & 98 & 2 & 28 \\
12 & 49 52 & 4 & 48 & 2 & 24 \\
10 & 55 & 3 & 30 & 1 & 10 \\
\hline 
Total & & 43 & 706 & 13 & 210 \\
\end{tabular}
\end{table}

\begin{figure}[ht]
    \includegraphics[width=8cm]{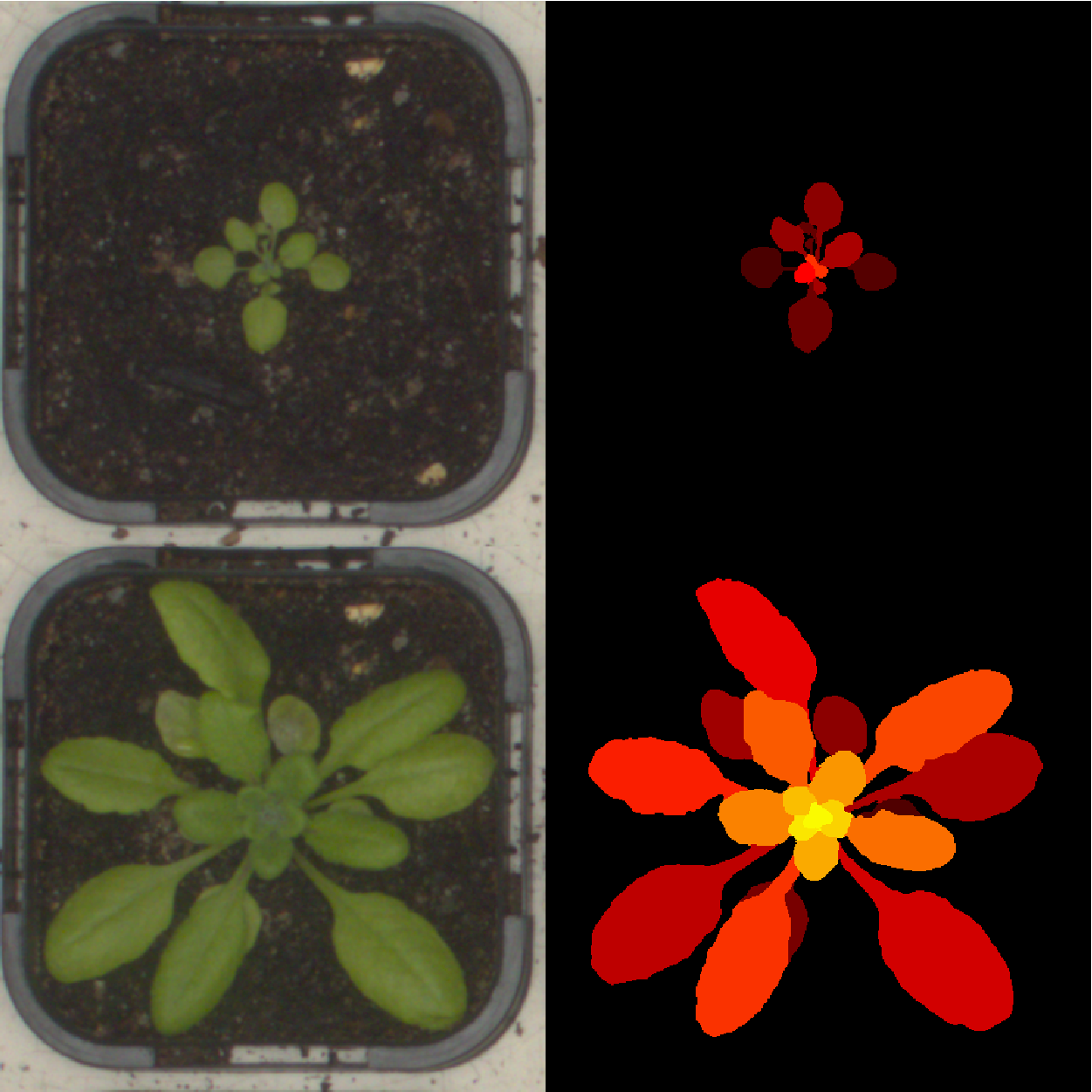}
    \caption{Two views of a plant from our dataset taken 27 and 42 days after sowing together with ground truth annotations.
    \label{fig:groundtruths}}
\end{figure}

The test set has no plants in common with the training data. 
The work described here used this test data / training data split. 
Examples of annotations from early and late in the growth
period are shown in Figure \ref{fig:groundtruths}.

\subsection{Evaluation of segmentation} \label{sec:dataset:eval}
Alongside the data and annotations we supply software for evaluation of leaf level 
segmentation that supports the 
subset-matched Jaccard index and an alternative implementation of symmetrical best Dice,
the evaluation measure used in the Leaf Segmentation challenge 
\cite{Scharr2014a,Minervini2014a}. There is a fuller description of the
subset-matched Jaccard index approach in \cite{Bell2016a}.
In this paper, we use Symmetric Best Dice for quantifying the quality of 
segmentations for the sake of compatibility with others' work, such as
\cite{Scharr2015a}.

\section{Method} \label{sec:method}

The starting point for this work was the empirical observation that segmentation
techniques based upon objects (for example, leaves) have difficulty distinguishing
individual leaves in the presence of heavy occlusion. For example, Yin et
al \cite{Yin2014a,Yin2014b} report an impressive technique for leaf 
segmentation, but acknowledge that it is less reliable in the presence of more 
than 25\% occlusion. With Arabidopsis plants, this level of occlusion will occur long 
before the plant nears maturity and flowering. For this reason we aim to detect
and classify edges in the image, determining whether an edge is an occluding edge or 
not, and to base our segmentation technique upon edge information rather than
upon models of the entire leaf.

There are further advantages to using an edge based approach. 
It can be expected that an effective edge based approach might outperform an object
based approach where there is on overlap leading to 
considerable occlusion of one leaf
as the shape of the overlapping pair of leaves will resemble a single leaf. 
An example of this appears in Figure \ref{fig:groundtruths}.
An edge based approach should also speed analysis over a sliding-window based 
technique, as the object
based approach will need to classify all pixels in an image while we need only 
analyse the edge pixels.  Finally, when it comes to machine learning methods, 
large amounts of training data are vital. One plant might provide 12 example 
leaves for a training algorithm, but hundreds of examples of edge pixels.  Thus by concentrating on a smaller part of the plant, we have a richer set of data to work with.

Given detected edges in an image, we also need to distinguish 
those edges that are internal to the plant from those that are the 
edges of the plant as a whole and those that occur in the 
background.  So we have a four way classification problem, summarised as follows:-
\begin{itemize}
\item \textbf{Background} - any edge wholly outside a plant
\item \textbf{Plant edge} - any edge where plant meets background - these delineate the silhouette
of the plant.
\item \textbf{Leaf edge} - any edge inside a plant that marks where two leaves overlap.
\item \textbf{Internal noise} - any edge that is inside a plant and is not an edge separating two 
overlapping leaves. These might be caused by shadows, specular reflections or venation of a leaf.
\end{itemize}

We found that conventional machine learning classifiers worked well on distinguishing plant from background but were
less successful at distinguishing between leaf edges
where one leaf overlaps another and spurious edges found within the area of a plant.
 While the main interest in this approach to 
classification is in distinguishing between these last two ``inside plant'' classes, 
our approach is very effective at classifying plant edges so is effective for plant from 
background segmentation.

The result of the edge classification process can be visualised as an image rather like a 
line drawing, as shown in the second element of Figure \ref{fig:image_process_steps}, with 
each class of edge depicted with a line of different colour. However, for obtaining measurements a
region based segmentation with each leaf a different value  
is preferable, which makes leaf area and leaf count easy to derive.  This requirement leads to the main disadvantage of our approach, as we need a second
post-classification step to convert our edge classifications to a regional segmentation, 
complicating the image processing sequence. It is necessary to:-

\begin{enumerate}
\item Locate image edges. This is done using the Canny edge detector. 
The only requirement is that sensitivity is high because as many 
of the plant and leaf edges should be identified as possible.
The classification stage identifies the irrelevant edges, thus overdetection is 
not important for us.
\item Classify edge pixels into one of the four classes outlined above. 
\item Convert the edge based segmentation into a region based one. This can be done using
conventional image processing techniques as described in Section \ref{sec:method:regions}
\end{enumerate}

The effectiveness of our final segmentation and leaf count thus depends not only on the
quality of edge classification, but also on how well the edge 
classifications can be translated to region based classifications.

\subsection{Edge classification} \label{sec:method:edges}

\begin{figure*}[ht]
    \centering
    \includegraphics[width=13cm,keepaspectratio=true]{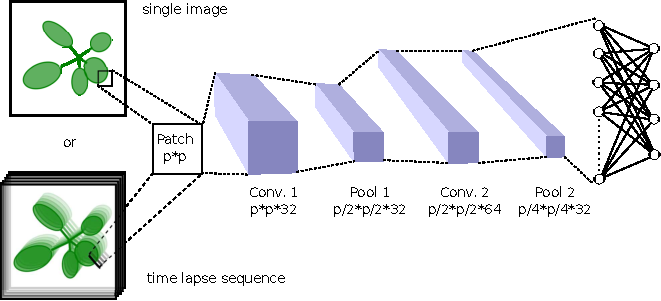}
    \caption{A simple deep learning network that alternates convolution 
    and pooling layers for four way classification of edge pixels.}
    \label{fig:CNN_layout}
\end{figure*}

Our classification techniques are based upon a convolutional 
neural network (CNN). Various configurations of this network were tried, and we 
report performance of the network under some of these configurations in Section~\ref{sec:results}.
The basic configuration is shown in Figure \ref{fig:CNN_layout}.

We have experimented with two key variations to the architecture. Firstly, we tried replacing the single four way classification by a tree of binary classifications
in which different networks (with identical architectures) are used to first separate
the internal edges from the others (i.e. plant edges and background) and then to subdivide each of those sets.
Secondly, we investigated ``temporal patches'', which build a patch to be classified from a line of pixels normal to the edge, taken from 
a sequence of time lapse images. The pixels from the same locations in each image are used. 
This is illustrated in the bottom of Figure \ref{fig:CNN_layout}.

Some further experimental configurations have been tested but did not improve the performance of our classifier, and so are not included in the results section. We mention them here for completeness. These were increasing the final pool output to 64, and doubling the pooling layers (where each convolutional layer is 
followed by a pair of pooling layers, making the sequence Conv. 1, Pool 1, Pool 2, 
Conv. 2, Pool 3, Pool 4 before the fully connected classification layer).

\subsection{Region based segmentation} \label{sec:method:regions}
\begin{figure*}[t!]
    \centering

    \subfloat[Remove isolated edges]{
        \includegraphics[height=1.2in]{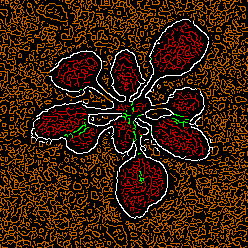}
    }
        \subfloat[Dilate internal edges]{
        \includegraphics[height=1.2in]{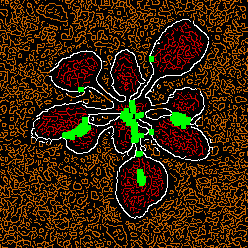}
    }
        \subfloat[Flood fill]{
        \includegraphics[height=1.2in]{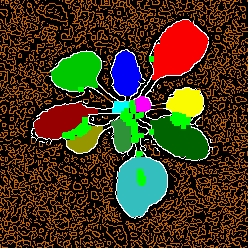}
    }
    
        \subfloat[Remove remaining edges]{
        \includegraphics[height=1.2in]{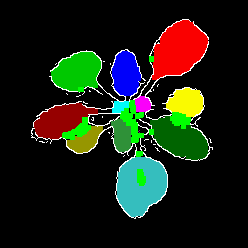}
    }
        \subfloat[Inflate leaf]{
        \includegraphics[height=1.2in]{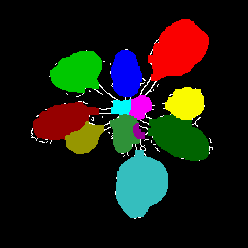}
    }
        \subfloat[Final noise removal]{
        \includegraphics[height=1.2in]{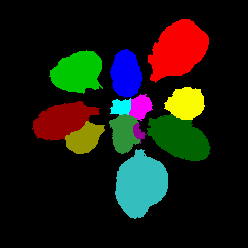}
    }
  
        \subfloat[Original Image]{
        \includegraphics[height=1.2in]{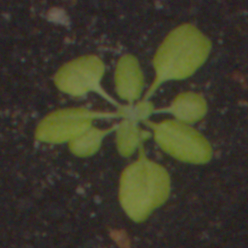}
    }
        \subfloat[Ground Truth]{
        \includegraphics[height=1.2in]{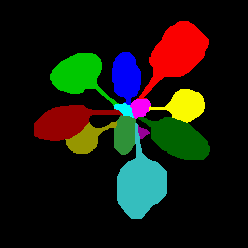}
    }
    \caption{The image processing steps from edge classification to the creation of a leaf-level segmented image.  Figures (a) to (f) correspond to steps (a) to (f) in the main text. \label{fig:image_process_steps}}
\end{figure*}

Once we have our edge classifications, this can be treated as an image consisting of lines
where each classification of edges is a line of a different colour. In the examples presented here,
plant edges are
white, leaf edges are green, background edges are orange and internal noise edges are red. 
However, the required result can be thought of as an image wherein each segmented region (leaf) is an area 
of a distinct colour, comparable to the ground truth images shown in \ref{fig:groundtruths}. In both cases, the background is black.
We use a sequence of conventional image processing techniques to convert our line
image to a region based segmented image from which measurements can be obtained by pixel 
counting. This is illustrated in Figure \ref{fig:image_process_steps}. 

The post classification step itself has seven steps, five of which are specific 
cases of dilation. These steps are:

\begin{enumerate}
\item Remove isolated "spots" of edge pixels classified as either plant edge or leaf edge.
Any regions of plant edge or leaf edge pixels that are isolated within a region of 
5 by 5 pixels are removed. If any of the pixels around the edge of this region are any 
colour other than the background colour, the pixels to be replaced are changed to 
that colour, otherwise they are changed to background colour.
\item Dilate the leaf edges. This reduces discontinuities resulting from the original edge 
detection step. In essence, the image as treated as binary with leaf edge pixels classed as 1 
and all other colours as 0, so any pixel that is within the range of dilation that is not leaf 
edge coloured is changed to that colour.  Again a 5 by 5 region is used.
\item Each distinct internal region (leaf) is coloured using flood fill.
Each individual leaf is assigned to its own class (that is, a different colour is used for each leaf). Flood fill is stopped when either leaf edge 
or plant edge pixels are encountered. A 5 by 5 structuring element is used to find a pixel that 
triggers the stopping of flood fill, so any remaining holes in edge pixel classifications do 
not result the flood fill ``leaking out''. This also prevents narrow petioles being filled.
\item Background edges are removed, replaced by background colour.
\item The flood filled leaf colours are dilated but only over leaf edge pixels. 
This continues until all leaf edge coloured pixels are replaced by a leaf colour. 
Leaf colours are dilated by one, but only over background coloured or plant 
edge coloured pixels. This would ideally result in the replacement of the plant edges but 
sometimes isolated pixels remain.
\item Finally, any remaining plant edge coloured pixels are changed to their neighbour's colour.
This will be any leaf colour that two neighbouring pixels share, or background colour 
if  there is no such colour.
\end{enumerate}
This results in a coloured image with a black background and each leaf (as detected by the 
classification) coloured a different colour, so it can be evaluated against the corresponding 
ground truth annotation to evaluate the quality of the segmentation. 
The results are discussed in the next section.

\section{Results} \label{sec:results}

Results are presented using the same evaluation measures as in the Leaf segmentation
Challenge, \cite{Scharr2014a,Minervini2014a}. These are 
Difference in Count (DiC), Absolute Difference in Count ($\vert$DiC$\vert$), Foreground/Background Dice (FBD), and Symmetric Best Dice (SBD).  With regard to DiC, figures quoted here are mean results from several plants, 
and so this measure is not very useful as over- and under- segmentation tend to cancel each other out. 
So the sign of this result here gives an indication of whether the approach tends to over or 
under segment, but it tells us little else.
The standard deviation (quoted in brackets in the tables)
is a more useful guide to accuracy in this case. The use of these metrics ensures our results can readily be compared with results in \cite{Scharr2015a}. 

Results presented here show the effects of varying aspects of our CNN setup relative to a ``baseline'' configuration. 
This baseline configuration is as follows:-

\begin{itemize}
\item Using a set of three binary classifications rather than one four way classification.
\item Using a patch from a single image, not one obtained from normals from a
timelapse sequence.
\item Using a patch size of 16 pixels square.
\item Using the CNN architecture illustrated in Figure \ref{fig:CNN_layout}
\end{itemize}

Experiments were run using training sets drawn from our dataset, and tested using the test sets as described earlier.  Results quoted are the mean of score for plants in one image of our test data set, the image being taken 31 days after sowing; this was chosen as it presents reasonable levels of occlusion and also corresponds closely to the data in the LSC challenge.  The results here presented omit those variations in network architecture which made little difference.  

\begin{table*}
\centering
\caption{The effect of CNN parameters quality of leaf counting and segmentation. The asterisked lines represent our baseline configuration, and standard deviations are presented in brackets}
\label{table:results_variation}
\begin{tabular}{l|c|c|c|c|c}
Experiment & Variant & DiC & $\vert$DiC$\vert$ & FBD & SBD \\

\hline
\multirow{4}{*}{Patch Size}& 8 & 5.94 (3.87) & 6.06 (3.69) & 0.94 (0.01) & 0.55 (0.12) \\
&12 & 3.94 (3.21) & 4.06 (3.06) & 0.95 (0.01) & 0.63 (0.11) \\
&16* & 1.06 (1.73) & 1.39 (1.46) & 0.95 (0.01) & 0.68 (0.09) \\
&32 & -0.33 (2.30) & 1.78 (1.44) & 0.95 (0.01) & 0.70 (0.08) \\

\hline
\multirow{2}{*}{Binary vs 4-way} & Binary* & 1.06 (1.73) & 1.39 (1.46) & 0.95 (0.01) & 0.68 (0.09) \\
& Four-way & -0.83 (2.31) & 2.06 (1.26) & 0.94 (0.01) & 0.71 (0.07) \\

\hline
\multirow{2}{*}{\parbox{4cm}{Single image vs temporal patch}} & Single*  & 1.06 (1.73) & 1.39 (1.46) & 0.95 (0.01) & 0.68 (0.09) \\
&Temporal & 3.78 (4.29) & 3.89 (4.19) & 0.95 (0.01) & 0.61 (0.12) \\
\hline
\end{tabular}
\end{table*}

Increasing the patch size up to 16 improved classification 
and segmentation. There was little if any further improvement increasing patch size
to 32. Increasing the patch size has the drawback that more edge pixels have their associated 
patches ``bleeding off'' the edge of the image.  In general, better results were obtained from using the binary sequence approach
rather than a single four way classification. 
Finally, the idea of making the patches from a time lapse sequence of images
did not reveal any improvement. It also introduces the problem of gaps in the sequence
because the use of visible light prevents images being taken at night,
leaving 12-hour gaps in the sequence each day. 

As Arabidopsis plants grow, the number of leaves increases as does the amount that they overlap 
so it might be expected that results for leaf counting and measurement will be worse with
older plants and Table \ref{table:results_older} demonstrates this.

\begin{table*}
\centering
\caption{The effect of using older plant (37 against 31 days after sowing) on quality of leaf counting and segmentation.}
\label{table:results_older}
\begin{tabular}{c|c|c|c|c}
Age of plants & DiC & $\vert$DiC$\vert$ & FBD & SBD \\
\hline
31 d.a.s. & 1.06 (1.73) & 1.39 (1.46) & 0.95 (0.01) & 0.68 (0.09) \\
37 d.a.s. & -0.94 (6.56) & 5.19 (3.90) & 0.96 (0.01) & 0.58 (0.10) \\
\cline{1-5}
\end{tabular}
\end{table*}

In addition to the experiments using our own images, some variations were tried 
with a publicly available subset of the Leaf Segmentation Challenge Dataset, the LSC ``A2'' dataset, that
being the set most similar to ours.
The motivation for doing this was to see how dependent our approach is to having
training data closely matched to the test data; as there are 31 ground-truthed plants in the LSC A2 dataset there were not enough instances to tailor our model by training our CNN from scratch. 
Thus our approach, trained on the 706 plants from the
ALED dataset, has a similar input to the LSC A2 dataset capture setup but is not a perfect match.
We could not try the ``temporal patches'' variations with this data owing to the
lack of time lapse data in the Leaf Segmentation Challenge dataset.  
In Table \ref{table:results_LSC_data} we present the baseline variations using our data and the same variations
using the LSC data as test date for comparison. 

\begin{table*}
\centering
\caption{Results on the LSC A2 dataset, compared with other techniques as reported in \cite{Scharr2015a}, \cite{Ren2017} and \cite{Romera-Paredes2016a}. Asterisked results are aggregated over all LSC datasets, rather than just L2.}
\label{table:results_LSC_data}
\begin{tabular}{l|c|c|c|c}
Method & DiC & $\vert$DiC$\vert$ & FBD & SBD \\
\hline
Our method & -0.3 (2.3) & 1.8 (1.4) & 0.95 (0.01) & 0.70 (0.08) \\
ITK  & -1.0 (1.5) & 1.2 (1.3)& 0.96 (0.01) & 0.77 (0.08) \\ 
Nottingham  & -1.9 (1.7)& 1.9 (1.7)& 0.93 (0.04) & 0.71 (0.1) \\ 
MSU  & -2.0 (1.5) & 2.0 (1.5) & 0.88 (0.36) & 0.66 (0.08)\\ 
Wageningen  & -0.2 (0.7) & 0.4 (0.5) & 0.95 (0.2) & 0.76 (0.08)\\ 
Romera-Paredes* & 0.2 (1.4) & 1.1 (0.9) & N/A & 0.57 (0.8)\\
Romera-Paredes+CRF* & 0.2 (1.4) & 1.1 (0.9) & N/A & 0.67 (0.9)\\
Ren* & N/A & 0.8 (1.0) & N/A & 0.85 (0.5) \\
\cline{1-5}
\end{tabular}
\end{table*}

\section{Discussion} \label{sec:discussion}
Our results are not dissimilar from those achieved by the approaches covered
in \cite{Scharr2015a}, and also those achieved by \cite{Romera-Paredes2016a} although a direct comparison is impossible due to differences in datasets. With the exception of \cite{Romera-Paredes2016a} all methods were trained on the LSC data, and ours was not.   Ren and Zemel, in \cite{Ren2017}, present considerably better results exceeding the current state-of-the-art on counting and symmetric best dice using a general purpose object counting network (RNN); parts of this system are pre-trained and this approach to learning shows a lot of promise.

While our approach does well at plant from background segmentation, 
there is still room for improvement in the leaf level results.  

In the hope of establishing to what extent our results are attributable to the use of
CNNs, we tried a Random Forest classifier (as implemented in Weka) to classify edges using patches. Table \ref{table:RandomForest} shows these results, and it is clear that
using a CNN gave a noticeable improvement in leaf count and symmetric best dice. These results
do raise the possibility of using a simpler classifier for plant from background segmentation
with a single CNN for classifying the two groups of edges found within the plant - overlapping 
leaf edges and others.

\begin{table*}
\centering
\caption{Comparison of results using the CNN and Random Forest on the same image with the same patch size (12).}
\label{table:RandomForest}
\begin{tabular}{c|c|c|c|c}
Classifier & DiC & $\vert$DiC$\vert$ & FBD & SBD \\
\hline
CNN & 1.4 (2.51) & 1.8 (2.17) & 0.95 (0.01) & 0.70 (0.13) \\
Random Forest & -2.4 (4.16) & 3.6 (2.88) & 0.95 (0.01) & 0.53 (0.13) \\
\cline{1-5}
\end{tabular}
\end{table*}

Most approaches using deep learning use objects rather than edges as the feature to be 
classified.  There seem to us to be advantages in using edges, 
at least for the problem we set out to solve. If the leaf is taken as the object to be 
classified, this is of variable shape and size, quite apart from the problem of
occlusion as the plant grows more leaves.
Many object based approaches appear to struggle once one of a pair of leaves is largely hidden.
As the percentage
of overlap increases, the overall shape of a pair of leaves more closely resembles
a possible single leaf shape, so an object based system might not detect the
existence of a second leaf, while the approach described here should.

When developing useful plant phenotyping systems, there are many considerations besides the range and accuracy of measurements obtainable from a system. We consider the criteria listed in \cite{Minervini2014a} an excellent set of characteristics for working in 
an interdisciplinary setting; in particular their emphasis on robustness and 
automation are key for high throughput work.  The approach described herein goes some way to meeting many these:

\begin{enumerate}
\item \emph{Accommodate most laboratory settings.} We were able to obtain results
on the LSC A2 dataset (which we had not trained the CNNs on) that approach the
results we obtained on our own dataset. This shows the approach is reasonably
flexible with respect to differences in set-up out of the box. 
\item \emph{Not rely on an explicit scene arrangement.} The method we have
developed is tailored to top-down views of rosette plants in trays, and we do not
expect it to work that well on other species or camera configurations. However
top-down imaging such as we analyse here is a very common setup for rosette plants.
\item \emph{Handle multiple plants in an image.} It does this, using high
throughput phenotyping images with up 20 plants in an image. 
\item \emph{Tolerate low resolution and out of focus discrepancies.} The camera
used was 5 Megapixels, by no means high by modern standards, and this resolution
was shared by up 20 plants. The camera was used \emph{as is}, with slight focus
discrepancies and barrel distortion.
\item \emph{Be robust to changes in the scene.}  This is not something we have
tested, given our robotic capture setup.
\item \emph{Require minimal user interaction, in both setting up and training.}
This is not clear at present. It depends on whether the CNNs could be trained on
images from several sources and give good performance on each source, which remains a direction for future work. 
\item \emph{Be scalable to large populations or high sampling rates.} It is likely
that this would depend on the computing power available, but as the technique is 
non-interactive, computer resource is the bottleneck (there is no human in the
loop).
\item \emph{Offer high accuracy and consistent performance.} This is good 
for plant from background segmentation but there is room for improvement at 
the leaf level, as discussed earlier. 
\item \emph{Be automated with high throughput.} 
Once trained the approach is entirely automated.
\end{enumerate}

\subsection{Future work} \label{sec:discussion:future}

One limitation of the work imposed by the nature of the available data is that the
CNNs were trained on data from the same capture setup as the test data.
This obvious issue is driven by the need to train the CNNs on a large, labelled 
dataset.  The
broader utility of the system would depend on how well a CNN trained on images 
from multiple sources would work. We show some positive results in the direction
of cross-dataset transferability (networks trained upon the ALED dataset
performing well on the LSC), but we
acknowledge that this is an area for future development.

We have made 
the datasets we used for this work publicly available and encourage 
other researchers in the field to do the same, as
sufficient publicly available datasets will be invaluable in 
advancing the state of the art. Given the dominance of CNN-based
imaging techniques in computer vision, and the reliance of such techniques on large quantities
of training data, \emph{the dataset question} is a becoming major one. The presence of more, larger datasets such as ALED can only help here.

With regard to deployment, it would be possible
to train the network as it stands as part of the installation process for
a phenotyping system. In this case,  
the installed system could in a sense be tuned to images
from a particular capture setup. This is clearly not ideal, especially 
as a large set of ground truth data 
would be needed, but might be a good solution for larger phenotyping
installations.

As we are not reliant on models of leaf shape, it seems reasonable to
expect that the
approach might be readily usable with other rosette plants beside Arabidopsis. It would be
interesting to test the approach against datasets of other plants and
especially against data sets of multiple plant species for both training and 
test data.
Again, lack of available data has prevented this at the current time, although this is certainly the direction of our future work plans. 

\section{Conclusion}\label{sec:conclusion}

The method of image analysis presented here appears to offer 
performance comparable to the 
current state of the art. Results for plant from background 
segmentation are encouraging
and our leaf counting results seem to offer an improvement over many of 
those in \cite{Scharr2015a}.

While there remains scope for improvement in both leaf 
counting and leaf segmentation performance we suggest that the 
approach offers a route towards further developments and that 
such developments 
will be both encouraged and enabled by the release of our data set 
and accompanying annotations to the wider community. We hope other 
researchers will do the same. The potential benefits of 
automated image analysis for phenotyping amply justify the 
continued development of work such as that presented here.

\section*{Availability of data and material}

The dataset is hosted at \url{https://zenodo.org} and it can be found from
\url{https://doi.org/10.5281/zenodo.168158}.

% if your bibliography is in bibtex format, use those commands:
\bibliographystyle{IEEEtran} % Style BST file (bmc-mathphys, vancouver, spbasic).
\bibliography{ArabidopsisLeafSegmentation}      % Bibliography file (usually '*.bib' )
% for author-year bibliography (bmc-mathphys or spbasic)
% a) write to bib file (bmc-mathphys only)
% @settings{label, options="nameyear"}
% b) uncomment next line
%\nocite{label}

% or include bibliography directly:
% \begin{thebibliography}
% \bibitem{b1}
% \end{thebibliography}

%%%%%%%%%%%%%%%%%%%%%%%%%%%%%%%%%%%
%%                               %%
%% Figures                       %%
%%                               %%
%% NB: this is for captions and  %%
%% Titles. All graphics must be  %%
%% submitted separately and NOT  %%
%% included in the Tex document  %%
%%                               %%
%%%%%%%%%%%%%%%%%%%%%%%%%%%%%%%%%%%

%%
%% Do not use \listoffigures as most will included as separate files

\end{document}